\documentclass[lettersize,journal]{IEEEtran}
\usepackage[T1]{fontenc}
\usepackage{amsmath,amsfonts}
\usepackage{algorithmic}
\usepackage{algorithm}
\usepackage{array}
\usepackage[caption=false,font=normalsize,labelfont=sf,textfont=sf]{subfig}
\usepackage{textcomp}
\usepackage{stfloats}
\usepackage{url}
\usepackage{verbatim}
\usepackage{graphicx}
\usepackage{cite}
\usepackage{booktabs}
\usepackage{xcolor}
\usepackage[table]{xcolor}
\begin{document}

\title{Toward Fine-Grained Facial Control in 3D Talking Head Generation}

\author{Shaoyang Xie, Xiaofeng Cong, Baosheng Yu, Zhipeng Gui, ~\IEEEmembership{Member},~IEEE, Jie Gui,~\IEEEmembership{Senior Member},~IEEE, Yuan Yan Tang,~\IEEEmembership{Life Fellow},~IEEE, James Tin-Yau Kwok,~\IEEEmembership{Fellow},~IEEE
\thanks{S. Xie and X. Cong are with the School of Cyber Science and Engineering, Southeast University, Nanjing 210000, China (e-mail: seuxsy@163.com, cxf\_svip@163.com). 
}
\thanks{B. Yu is with the Lee Kong Chian School of Medicine, Nanyang Technological University, 308232, Singapore (e-mail: baosheng.yu@ntu.edu.sg).}

\thanks{Z. Gui is with the School of Remote Sensing and Information Engineering, Wuhan University, Wuhan 430079, China, and also with the Collaborative Innovation Center of Geospatial Technology, Wuhan University, Wuhan 430079, China. (e-mail: zhipeng.gui@whu.edu.cn)}

\thanks{J. Gui is with the School of Cyber Science and Engineering, Southeast University and with Purple Mountain Laboratories, Nanjing,  and with Engineering Research Center of Blockchain Application, Supervision And Management (Southeast University), Ministry of Education, 210000, China (e-mail: guijie@seu.edu.cn).}

\thanks{Y. Tang is with the Department of Computer and Information Science, University of Macau, Macau 999078, China (e-mail: yytang@um.edu.mo).}

\thanks{J. Kwok is with the Department of Computer Science and Engineering, The Hong Kong University of Science and Technology, Hong Kong 999077, China (e-mail: jamesk@cse.ust.hk).}


}

\markboth{Journal of \LaTeX\ Class Files,~Vol.~14, No.~8, August~2021}%
{Shell \MakeLowercase{\textit{et al.}}: A Sample Article Using IEEEtran.cls for IEEE Journals}


\maketitle

\begin{abstract}
Audio-driven talking head generation is a core component of digital avatars, and 3D Gaussian Splatting has shown strong performance in real-time rendering of high-fidelity talking heads. However, achieving precise control over fine-grained facial movements remains a significant challenge, particularly due to lip-synchronization inaccuracies and facial jitter, both of which can contribute to the uncanny valley effect. 
To address these challenges, we propose Fine-Grained 3D Gaussian Splatting (FG-3DGS), a novel framework that enables temporally consistent and high-fidelity talking head generation. Our method introduces a frequency-aware disentanglement strategy to explicitly model facial regions based on their motion characteristics. Low-frequency regions, such as the cheeks, nose, and forehead, are jointly modeled using a standard MLP, while high-frequency regions, including the eyes and mouth, are captured separately using a dedicated network guided by facial area masks. The predicted motion dynamics, represented as Gaussian deltas, are applied to the static Gaussians to generate the final head frames, which are rendered via a rasterizer using frame-specific camera parameters. Additionally, a high-frequency-refined post-rendering alignment mechanism, learned from large-scale audio–video pairs by a pretrained model, is incorporated to enhance per-frame generation and achieve more accurate lip synchronization.
Extensive experiments on widely used datasets for talking head generation demonstrate that our method outperforms recent state-of-the-art approaches in producing high-fidelity, lip-synced talking head videos.
\end{abstract}

\begin{IEEEkeywords}
Talking head generation, 3D Gaussian Splatting.
\end{IEEEkeywords}

\section{Introduction}
\IEEEPARstart{A}{udio}-driven 3D talking head generation is emerging as a transformative technology at the forefront of digital human synthesis, with far-reaching applications in digital avatars~\cite{thies2020neural}, film and content production~\cite{zhang2021flow}, and real-time interactive communication~\cite{kim2018deep}. Its main research goal is to use algorithms to produce high-fidelity, lip-synchronized facial animations from speech signals~\cite{sheng2023toward,li2026cofaco}, which is crucial for creating believable, emotionally expressive digital humans. However, achieving this requires a delicate balance: accurately modeling speech-driven facial dynamics while preserving a consistent and identity-faithful appearance remains a major challenge for the field.

Generative Adversarial Networks (GANs)~\cite{goodfellow2020generative} have been extensively applied to talking head generation~\cite{guan2023stylesync, wang2023seeing, zhong2023identity}, achieving better lip-audio synchronization by leveraging large-scale audio-visual datasets~\cite{prajwal2020lip, zhang2023dinet}. Despite these advances, GAN-based methods often struggle to preserve speaker identity and produce high-resolution outputs. To overcome these limitations, recent studies have shown that incorporating 3D geometric priors is crucial~\cite{ji2022eamm, xing2023codetalker, zhang2022sadtalker, ren2021pirenderer, zhang2021flow, fan2022faceformer, peng2025synctalk++}. Inspired by the 3D Morphable Model (3DMM)~\cite{blanz2023morphable}, recent methods such as SadTalker~\cite{zhang2022sadtalker} address these challenges by predicting 3DMM coefficients for pose and expression directly from audio. While this strategy enhances identity preservation and ensures geometric consistency, it often results in animations that appear stiff or lack emotional expressiveness, due to the limited representational power of the linear 3DMM basis.

\begin{figure}[t]
\begin{center}
\includegraphics[width=1\linewidth]{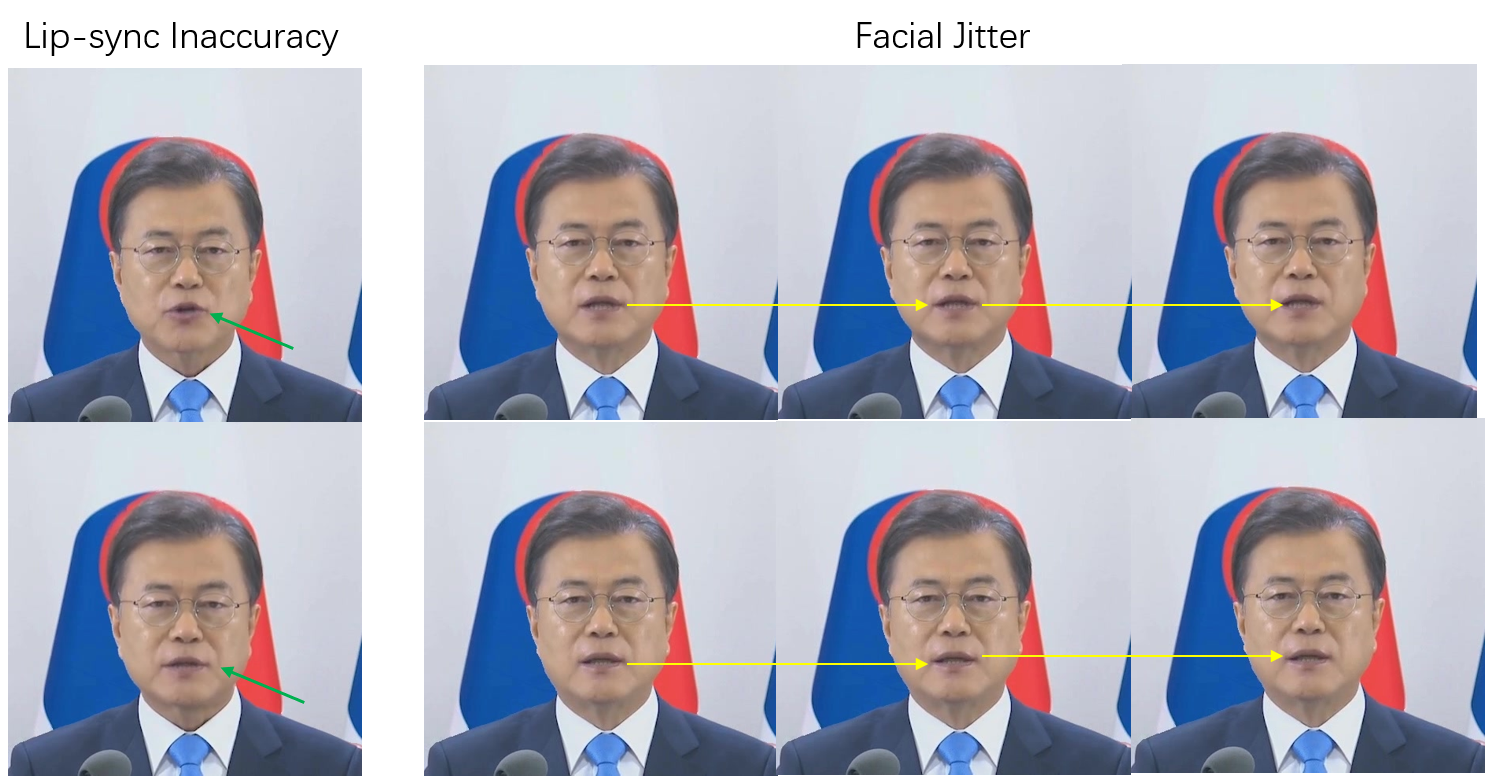}
\end{center}
\caption{\textbf{Illustration of lip-synchronization errors and facial jitter.} The top row shows generated frames, while the bottom row shows the corresponding ground-truth frames. Arrows highlight noticeable discrepancies, demonstrating that existing methods suffer from lip-synchronization inaccuracies and unstable facial motions.}
\label{fig:motivation}
\end{figure}

The success of NeRF~\cite{mildenhall2021nerf,huang2025nerf,chen2024learning} in novel view synthesis has significantly advanced the generation of high-fidelity, 3D-consistent talking heads~\cite{guo2021ad,tang2022real,li2023efficient,shen2023sd,wang2024high}. However, NeRF-based approaches are often limited by slow training and rendering speeds, making them less practical for real-time applications. To overcome these limitations, 3DGS~\cite{kerbl3Dgaussians,bao2025loopsparsegs,wang2025plgs} has recently emerged as a powerful alternative, offering both high-quality rendering and real-time performance. More importantly, 3DGS provides an explicit spatial representation, enabling finer control over motion and appearance, which is an essential capability for talking head generation~\cite{qian2024gaussianavatars,cho2024gaussiantalker,li2024talkinggaussian,li2025instag,gong2025monocular,shen2025audio,aneja2025gaussianspeech,bao20253d,ma2026esgaussianface}. For example, GaussianTalker~\cite{cho2024gaussiantalker} animates a canonical set of Gaussians using audio-driven deformations, achieving efficient and expressive results. Similarly, TalkingGaussian~\cite{li2024talkinggaussian} employs a motion prediction network to estimate point-wise offsets for each Gaussian, facilitating dynamic and accurate head animation. Though the above methods exhibit strong overall consistency, they often struggle to accurately capture fine-grained facial motions, especially in high-frequency regions like the mouth and eyes, which change rapidly during speech. As illustrated in Fig.~\ref{fig:motivation}, these subtle inaccuracies can lead to an uncanny valley effect, diminishing the perceived realism of the generated avatars.

In this paper, we propose Fine-Grained 3D Gaussian Splatting (FG-3DGS) to address the limitations of existing audio-driven talking head generation methods, particularly their difficulty in modeling fine-grained facial motions. Facial regions exhibit heterogeneous motion patterns: high-frequency areas such as the eyes and mouth undergo rapid, complex movements (e.g., lip-synchronization changes and blinking), while low-frequency regions like the cheeks, forehead, and jaw primarily follow rigid head-pose movements and exhibit relatively minor changes. Moreover, simply employing a unified model to capture motion across all regions often fails to effectively capture these disparate dynamics, leading to biased training toward the dominant low-frequency areas. Consequently, these models tend to prioritize appearance over accurate motion representation, leading to unsmooth results, particularly at the boundaries between high- and low-frequency zones. To overcome this issue, FG-3DGS introduces a frequency-aware disentanglement strategy that explicitly separates the modeling of high- and low-frequency regions. For high-frequency regions, we use specialized subnetworks that independently predict Gaussian deltas, enabling precise animation of the eyes and mouth. In contrast, a shared lightweight MLP is used to model the more rigid motions in low-frequency regions. This design enables region-specific learning of motion dynamics from audio signals and is conceptually aligned with the slow-fast network paradigm used in spatiotemporal modeling~\cite{feichtenhofer2019slowfast}. Furthermore, leveraging the differentiability of Gaussian splatting rasterization, we incorporate a high-frequency-refined post-rendering alignment mechanism to guide the training process using consecutive frames, thereby enhancing temporal coherence and improving lip synchronization. By combining frequency-specific branches based on facial region masks, our method produces expressive, temporally consistent talking head animations. Our main contributions are summarized as follows:
\begin{itemize}
    \item We propose \textbf{FG-3DGS}, a novel framework for talking-head generation that effectively captures heterogeneous motion across facial regions.
    \item We introduce a \textbf{frequency-aware disentanglement strategy} that separately models low- and high-frequency regions, mitigating the dominance of appearance-based gradients in highly dynamic areas.
    \item We develop a \textbf{high-frequency-refined post-rendering alignment} mechanism that utilizes a pre-trained lip-synchronization discriminator on the rendered outputs. This ensures precise synchronization between audio and high-frequency lip movements through fine-grained supervision.
\end{itemize}

\section{Related Work}
\subsection{Audio-Driven Talking Head Generation}
Audio-driven talking head generation aims to produce videos of a target person that satisfy the dual goals of photorealism and precise audio-visual synchronization. Early methods primarily rely on 2D GANs to generate audio-synchronized mouth movements from a single image or video. For example, Wav2Lip~\cite{prajwal2020lip} employs pre-trained lip-synchronization discriminators to align mouth movements with speech, achieving high synchronization quality. Although these approaches produce plausible results under constrained viewpoints, they struggle to generalize to dynamic head poses due to the inherent limitations of 2D representations. Subsequent efforts address these geometric limitations by integrating intermediate 3D priors. For instance, several 3D model-based methods~\cite{thies2020neural,wang2020mead,lu2021live,zhang2022sadtalker} utilize facial landmarks and 3DMM to disentangle pose and expression parameters, enabling more precise control of talking heads. While these pipelines improve pose controllability, their reliance on error-prone intermediate estimates, such as landmark detection inaccuracies, often compromises identity preservation, especially under extreme head rotations or challenging lighting conditions.

\subsection{NeRF-Based Talking Head Generation}
NeRFs have substantially advanced audio-driven 3D talking head generation by enabling the representation of detailed 3D facial structures. Early NeRF-based methods, such as AD-NeRF~\cite{guo2021ad}, demonstrate the viability of audio-conditioned NeRF for high-quality facial animation via person-specific training. However, vanilla NeRF suffers from slow rendering speeds and high computational costs, limiting its practical application. Subsequent efforts focus on improving rendering efficiency while maintaining high quality. RAD-NeRF~\cite{tang2022real} first introduces grid-based NeRF to accelerate computation. ER-NeRF~\cite{li2023efficient} reformulates the standard 3D hash grid into a tri-plane hash representation~\cite{chan2022efficient}, which reduces hash collisions and achieves a better balance between quality and efficiency. To enhance multimodal fusion, GeneFace~\cite{ye2023geneface} and SyncTalk~\cite{peng2024synctalk} leverage large-scale audio-visual datasets to pre-train dedicated audio encoders before NeRF rendering. Although NeRF-based methods achieve high reconstruction fidelity and multi-view consistency, addressing the entangled representations of rigid craniofacial geometry and achieving faster rendering speeds remain ongoing challenges.

\begin{figure*}[t]
\centering
\includegraphics[width=1\textwidth]{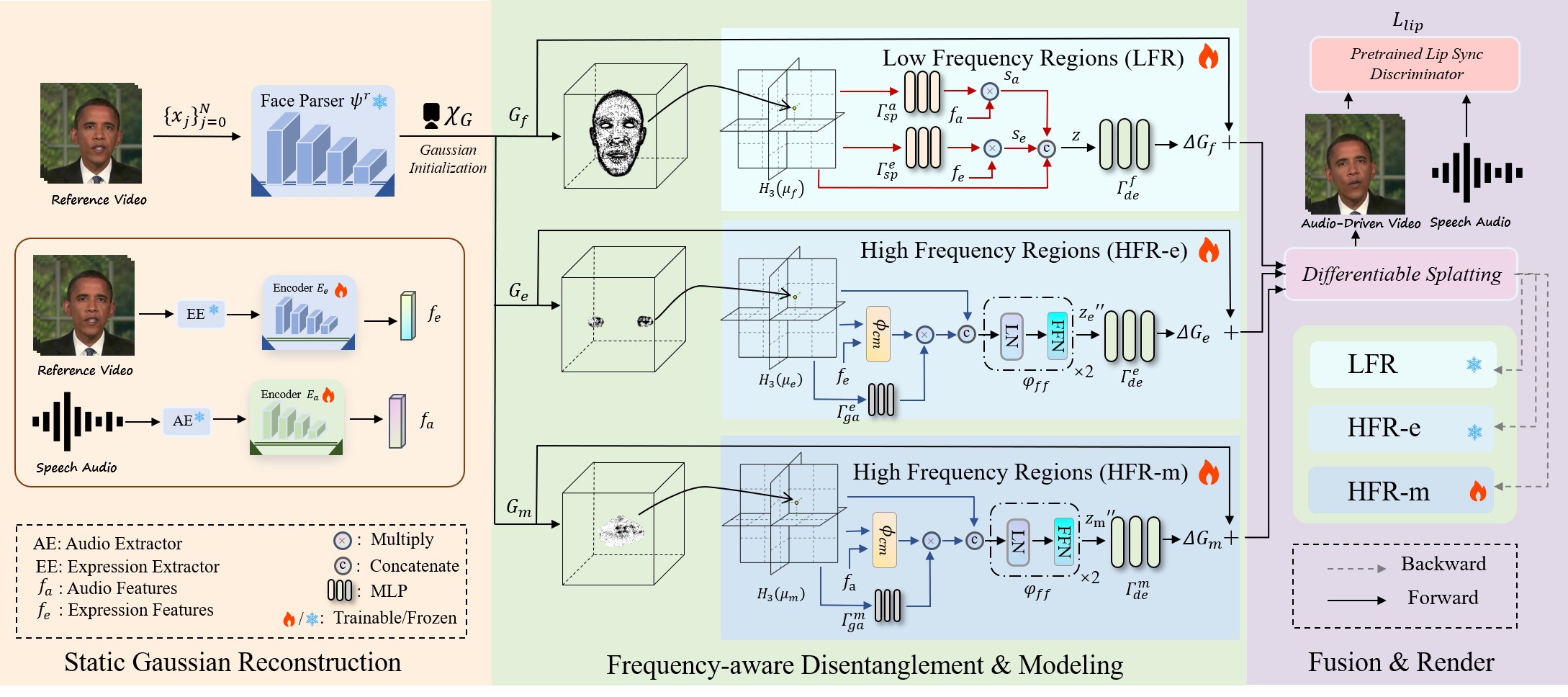}
\caption{\textbf{The main proposed FG-3DGS framework for 3D talking head generation}. Given a specific portrait speech video, FG-3DGS first decomposes the head into three regions: the face, eyes, and mouth. After performing static 3D Gaussian reconstruction, a conditional deformation attention mechanism predicts Gaussian offsets based on the encoded audio feature $f_a$ and expression feature $f_e$. The outputs from the static and dynamic components are then combined, and a 3D Gaussian rasterizer renders the dynamic Gaussians into images under varying camera parameters. To enhance lip synchronization, the high-frequency refined post-rendering alignment is applied in the final stage.}
\label{fig:the_overall_framework}
\end{figure*}

\subsection{3DGS-Based Talking Head Generation}
Unlike NeRF, which represents 3D structures implicitly through neural networks, 3DGS~\cite{kerbl3Dgaussians} uses an explicit point cloud representation to construct the radiance field. This approach achieves much faster rendering and higher-quality results for multi-view static scene reconstruction. Although 3DGS was originally developed for static scenes, dynamic variants such as 4DGS~\cite{wu20244d} and Dynamic3DGS~\cite{luiten2024dynamic} have extended its use to reconstructing dynamic 3D scenes. GSTalker~\cite{chen2024gstalker} is the first to apply deformable Gaussian splatting to audio-driven talking head generation, opening new possibilities for dynamic 3DGS applications. Building on this, GaussianTalker~\cite{cho2024gaussiantalker} employs a cross-modal attention mechanism to better fuse audio and spatial features, resulting in vivid and expressive facial animations. Additionally, TalkingGaussian~\cite{li2024talkinggaussian} improves fine-grained speech motion by separating the modeling of intra-oral regions from other facial areas. However, current methods often ignore the role of eye movements, which are crucial for identity preservation and emotional expression. Likewise, existing 3DGS approaches often overlook lip-synchronization accuracy, leading to unnatural mouth motion.

Generating realistic talking heads hinges not just on visual quality, but on the faithful reproduction of complex, non-uniform facial dynamics. Although previous NeRF-based methods\cite{liu2022semantic,li2023efficient,guo2021ad,shen2022learning,tang2022real} and 3D Gaussian-based methods\cite{wang2024gaussianhead,chen2024gstalker,cho2024gaussiantalker} have made a great contribution to the high-quality talking head synthesis, they treat the talking head as a single integral entity, learning a direct mapping from audio signals to complete facial representation. These approaches often yield overly-smoothed or averaged facial dynamics, which contradict the true biomechanics of the human face: different facial regions move in an unsmooth manner and exhibit distinct motion characteristics. This often results in a rigid coupling of facial regions, leading to the uncanny valley effect.

\section{Method}
In this section, we introduce the proposed FG-3DGS framework for talking head generation. We first describe the construction of static Gaussians, following the previous method~\cite{kerbl3Dgaussians}. After that, we describe the frequency-aware disentanglement and modeling of different regions. Next, we introduce a high-frequency-refined post-rendering alignment mechanism to synchronize lip movement and audio under supervision. Fig.~\ref{fig:the_overall_framework} illustrates the main proposed FG-3DGS framework for talking head generation.

\subsection{Static Gaussian Reconstruction}
To disentangle the frequency-aware attributes within the 3D Gaussian space, we reconstruct a static head Gaussian model in the canonical space by 3D Gaussian Splatting (3DGS) \cite{kerbl3Dgaussians}. A collection of 3D Gaussians $\mathcal{G} = \{G_1, G_2, \dots, G_{i}, \dots \}$ is employed to represent 3D structures, where $i$ denotes the index. Each Gaussian primitive is defined by a set of optimizable attributes, which include a 3D mean position $\boldsymbol{\mu}_i \in \mathbb{R}^3$, a positive semi-definite covariance matrix $\boldsymbol{\Sigma}_i \in \mathbb{R}^{3 \times 3}$, an opacity value $\alpha_i \in \mathbb{R}$, and its color represented by Spherical Harmonics (SH) coefficients $\mathbf{c}_i \in \mathbb{R}^{3 \times 16}$. To ensure positive semi-definiteness and to control the rotation and scaling of each 3D Gaussian primitive, its covariance matrix is decomposed into a scaling matrix $S$ and a rotation matrix $R$ by
\begin{equation}
        \Sigma=RSS^{\top}R^{\top},
\end{equation}
where $R$ and $S$ are defined by a scaling factor $s \in \mathbb{R}^3$ and a rotation quaternion $r \in \mathbb{R}^4$, respectively. Therefore, the complete parameters of the i-th Gaussian primitive ${{G}}_i$ is defined as
\begin{equation}
        {G}_i=\{\mu_i,r_i,s_i,\mathbf{c}_i,\alpha_i\},
\end{equation}

Following the above procedure, a complete static 3D Gaussian representation $\mathcal{G}=\{\boldsymbol\mu,\mathbf{r},\mathbf{s},\mathbf{c},\boldsymbol\alpha\}$ can be constructed. Then, we can render a coarse static head of a specific person from a few-minute portrait video. Sharing the similar problem settings with previous NeRF-based methods \cite{guo2021ad,li2023efficient,tang2022real}, we track the head pose and inversely calculate the camera pose $\{\pi_i\}_{i=1}^N$ from a few-minute portrait video $V=\{I_n\}_{n=1}^N$ of a specific person to reconstruct different scene, where $N$ represents the length of frames. The pixel-wise L1 loss and the D-SSIM loss \cite{wang2004image} $\mathcal{L}_{DS}$ are employed to measure the difference between the mask ground-truth image $x^{ms}$ and the static rendered image $x^{st}$. The loss function for the static Gaussian reconstruction is
\begin{equation}
   \mathcal{L}_{st} = \mathcal{L}_1(x^{ms}, x^{st}) + \lambda_{1} \mathcal{L}_{DS}(x^{ms}, x^{st}),
\end{equation}
where $\lambda_{1}$ denotes the weight factors. Next, the static Gaussians $G_r$ are jointly optimized with the frequency-aware modeling.

\subsection{Frequency-Aware Disentanglement}
Previous methods~\cite{wang2024gaussianhead,chen2024gstalker,cho2024gaussiantalker} have made an impressive contribution to the high-quality talking head synthesis. They treat the talking head as a single integral entity, but this often results in a rigid coupling of facial regions, leading to the uncanny valley effect. In contrast to the previous method, we leverage the fine-grained facial region disentanglement, modeling the talking head as a collection of regions with different frequency characteristics namely: (i) the mouth region $\mathcal{G}_{m}$ and eye region $\mathcal{G}_{e}$ with high-frequency movements that are tightly correlated with the input audio, and (ii) the face region $\mathcal{G}_{f}$ with low-frequency exhibit relatively minor changes driven a combination of both audio input and expression signals suggested by the upper face movement. 

To parse the talking head into separate parts, we used the settings from TalkingGaussian. A semantic mask generated by two face parsing models is used to divide the entire face into distinct regions. First, the BiSeNet~\cite{yu2018bisenet} parser, pretrained on the CelebAMask-HQ dataset~\cite{lee2020maskgan}, is used to predict a coarse mask of the high-frequency regions: eyes and mouth. Given the possible domain gap introduced by the parser, it may fail to cover the entire mouth in some `closed' cases. The other face parse model, a ResNet-based FPN pretrained on Easyportrait~\cite{kvanchiani2023easyportrait}, is introduced to generate the tooth mask. Then the two masks are overlaid to obtain the finer one. $\Psi_{x_{j}}^{r}$ denotes the mask for the $j$-th frame $x_{j}$ in the $r$-th region. The static Gaussian for each region can be obtained by
\begin{equation}
    \mathcal{G}_{r} = \Phi_{G}(\{\Psi_{x_{0}}^{r} \odot x_{0},..., \Psi_{x_{j}}^{r} \odot x_{j}\}, \{\pi_{0},...,\pi_{j}\}),
\end{equation}
where $r \in \{f, m, e\}$, $\Phi_{G}$ represents the Gaussian initialization process. The $\odot$ denotes the pixel-wise product.

\subsection{Frequency-Aware Modeling}

After establishing the static Gaussian head $\mathcal{G}_{r}$, we aim to learn the connection between the input audio $a$ and the motion offset $\Delta G_{r}$ of the three regions $G_r$. Recognizing that different facial regions exhibit different frequency characteristics, we employ a jointly emotion- and audio-driven motion prediction network for the facial region $\mathcal{G}_{f}$ and a gated cross-modal motion prediction network for the eye region $\mathcal{G}_{e}$ and mouth region $\mathcal{G}_{m}$. The efficient triplane plane hash encoder \cite{li2023efficient,li2024talkinggaussian} $\mathcal{H}_3$ is adopted to encode the three-dimensional 3D Gaussian position $\mu$ into a multi-resolution representation. The modeling process for low-frequency and high-frequency regions is as follows.\\

\noindent \textbf{Low-Frequency Region.} Typically, the movements associated with the low-frequency region represent coarse-grained structural dynamics that remain relatively invariant to transient speech signals or subtle, rapid head repositioning. Given this inherent stability, employing a high-capacity or overly complex motion prediction network may introduce the risk of overfitting, leading the model to capture idiosyncratic noise rather than meaningful motion patterns. To mitigate this, we propose a streamlined, lightweight architecture that achieves an optimal balance between expressive sufficiency and computational efficiency. Specifically, for the 3D Gaussians residing within this region, we predict their corresponding motion offsets through a jointly emotion- and audio-driven framework. This framework is underpinned by a multi-layer perceptron (MLP) network that integrates a regional attention mechanism with a strategic feature concatenation approach. To initiate the process, we leverage the robust representative power of pre-trained models. Specifically, we adopt an audio-speech extractor \cite{hannun2014deep,hsu2021hubert} and a dedicated emotion extractor to derive raw frequency-domain information $a$ and emotional latent information $e$, respectively. As illustrated in Fig.~\ref{fig:the_overall_framework}, these raw inputs $a$ and $e$ are further transformed by learnable encoding networks $\tau_{a}$ and $\tau_{e}$ to generate refined audio features $f_a$ and expression-rich features $f_e$. This encoding process is formally defined as
\begin{equation}
    \mathbf{f}_a = \tau_{a}(a), \mathbf{f}_e = \tau_{e}(e).
\end{equation}

To incorporate spatial context, we utilize the encoded 3D Gaussian positions $h_{f}=\mathcal H_{3}(\mu_{f})$ derived from the hash encoding. These spatial features are processed by dedicated MLPs to generate region-specific modulation weights
\begin{equation}
    \mathbf{s}_{a} = \Gamma_{sp}^{a}(h_{f}), \mathbf{s}_{e} = \Gamma_{sp}^{e}(h_{f}),
\end{equation}
where $\Gamma_{sp}^{a}$ and $\Gamma_{sp}^{e}$ denote the spatial MLPs for audio and expression modalities, respectively. The resulting spatial weights, $\mathbf{s}_{a}$ and $\mathbf{s}_{e}$, act as an attention mechanism, determining the localized influence of audio and emotional signals at different coordinates within the residual facial region. By applying this spatial gating, the model can adaptively prioritize relevant features for motion synthesis. The final integrated feature representation $\mathbf{z}$, which combines the spatial backbone with the modulated audio and emotion streams, is obtained through an adaptive fusion process:
\begin{equation}
    \mathbf{z} = [\mathcal H_{3}(\mu_f) \oplus(\mathbf{s}_{a} \odot \mathbf{f}_a) \oplus (\mathbf{s}_{e} \odot \mathbf{f}_e)],
\end{equation}
where $\oplus$ represents the concatenation operator and $\odot$ denotes the element-wise Hadamard product. This comprehensive representation $\mathbf{z}$ is subsequently fed into a deformation MLP, $\Gamma_{\text{de}}^f$, which is optimized to regress the dynamic offsets for the Gaussian attributes. This final step is formulated as
\begin{equation}
    \Delta G_f = \Gamma_{\text{de}}^f(\mathbf{z}),
\end{equation}
where $\Delta G_f = \{\Delta\mu_f,\Delta s_f,\Delta r_f\}$ encompasses the predicted motion offsets for the 3D Gaussian parameters, including position, scale, and rotation, thereby enabling realistic and synchronized facial dynamics. \\

\noindent\textbf{High-Frequency Regions.} In contrast to the relatively stable facial areas, the motions of the mouth and eye regions are characterized by high-frequency dynamics and intricate local deformations. In these regions, the mouth movements are primarily and directly driven by temporal audio signals, while eye movements are predominantly governed by expression latent signals. Due to the non-linear complexity of these motions, a simple linear or shallow network fails to accurately model such rapid transitions, often leading to visual artifacts such as motion jitter, stiff transitions, or muted expressions that lack emotional depth. To effectively bridge and fuse multi-resolution spatial geometry with heterogeneous conditional information for these highly expressive regions, we introduce a gated cross-modal motion prediction network. The process begins by extracting localized geometric context. The spatial features of the mouth and eye regions are obtained by
\begin{equation}
    h_{m} = \mathcal{H}_{3}(\mu_{m}), \quad h_{e}=\mathcal{H}_{3}(\mu_{e}).
\end{equation}

To achieve precise synchronization, we merge the canonical 3D Gaussians with dynamic conditional features through a cross-attention mechanism: audio features $\mathbf{f}_{a}$ drive the rapid mouth movements, while expression features $\mathbf{f}_{e}$ control the subtle eye motions. This integration effectively captures how varying input conditions influence the underlying Gaussian motion patterns across different facial topologies. Furthermore, a lightweight gating MLP, $\Gamma_{ga}$, is adopted, which takes the spatial context ($\mathbf{f}_{a}$ or $\mathbf{f}_{e}$) as input to predict a learnable scalar gating value. This formulation offers a profound level of fine-grained control: the cross-modal attention block spatially aligns temporal driving signals with static facial geometry, while the parallel spatial gate dynamically filters and scales the intensity of motion based on the specific geometric location where it occurs. The scalar gating value for adaptive modulation is obtained by
\begin{equation}
    \lambda_{r} = \sigma (\Gamma_{ga}^{r}(\mathbf{f}_r)), \quad \mathbf{f}_{r} \in {\mathbf{f}_{a}, \mathbf{f}_{e}},
\end{equation}
where $\sigma$ is the sigmoid function, constraining $\lambda_{r} \in (0, 1)$, with $r$ denoting the region index. The cross-modal prediction network consists of a cross-modal attention block, $\phi_{cm}$, and multiple feed-forward layers, $\varphi_{ff}$, interconnected via skip connections to ensure stable gradient flow. The computational pipeline for the prediction network is defined as:
\begin{equation}
    \begin{split}
        &\mathbf{z}_r = \lambda_r \odot\phi_{cm}(\mathbf{f}_{r}, \mathbf{c}_{r}), \\
        &\mathbf{z}_r' = \varphi{'}_{ff}(\mathbf{f}_{r}+\mathbf{z}_r),\\
        &\mathbf{z}_r'' = \varphi{''}_{ff}(\mathbf{z}_r'+\mathbf{z}_r).
    \end{split}
\end{equation}

In this workflow, the condition and spatial features are first fused in the cross-modal attention block. Then, the resulting fused feature $\mathbf{z}_r$ is modulated element-wise by the spatial gate and processed through the feed-forward network (FFN) with dual residual connections to iteratively refine and extract the final motion information $\mathbf{z}_r''$. This final fused feature $\mathbf{z}_r''$ is ultimately passed to the deformation MLP $\Gamma_{de}^r$, which is formulated as
\begin{equation}
    \Delta G_{r} = \Gamma_{\text{de}}^r(\mathbf{z}_{r}'')={\Delta\mu_r},
\end{equation}where $\Delta G_{r} \in \{\Delta G_{m}, \Delta G_{e}\}$ represents the motion offset $\Delta\mu_r$ of the 3D Gaussian parameters, specifically targeting the displacement of means to capture high-frequency facial articulations. The loss function of the overall Frequency-Aware Modeling stage is formulated as:
\begin{equation}
   \mathcal{L}_{FM} = \mathcal{L}_1(x^{ms}, x^{dy}) + \lambda_{1} \mathcal{L}_{DS}(x^{ms}, x^{dy}),
\end{equation}
where $x^{dy}$ represents the dynamic rendered face.\\

\noindent \textbf{Fuse \& Render.}
To synthesize the final talking head sequence, the discrete renderings from the previously defined low-frequency and high-frequency regions must be seamlessly integrated into a unified image space. We implement the alpha-blending principles widely used in image processing to aggregate the regional outputs. This fusion process is formulated as:
\begin{equation}
    \begin{aligned}
        \mathcal{C}_{fuse} = & [(1-\alpha_m) \times (\mathcal{C}_{e}+\Delta\mathcal{C}{_e})+ \alpha_m \times (\mathcal{C}{_m}+\Delta\mathcal{C}{_m})] \\
        & \times (1-\alpha_f) + (\mathcal{C}{_f}+\Delta\mathcal{C}{_f}) \times \alpha_f,
    \end{aligned}
\end{equation}
where $\alpha_f$ and $\{\alpha_e, \alpha_{m}\}$ are the opacity from the low-frequency and high-frequency regions, $C$ represents the pixel color. During the fusion stage, apart from the reconstruction loss, we randomly cut patches from the images and manipulate the LPIPS loss $\mathcal{L}_{LP}$ \cite{zhang2018unreasonable,li2023efficient,li2024talkinggaussian} to improve the details of the generated images. The overall loss function between the fusion and rendered image $x^{fu}$ and $x$ is
\begin{equation}
\begin{aligned} 
        \mathcal{L}_{fu} = \mathcal{L}_1(x, x^{fu}) + \lambda_{1} \mathcal{L}_{DS}(x, x^{fu}) + \lambda_{2} \mathcal{L}_{LP}(x, x^{fu}), 
\end{aligned}
\end{equation}
where $\lambda_{2}$ is the weight factor.

\subsection{High-Frequency-Refined Post-Rendering Alignment}
To enhance lip synchronization in 3D Gaussian reconstruction, we introduce a lip-synchronization discriminator~\cite{prajwal2020lip} to compute the lip-synchronization loss $\mathcal{L}_{lip}$. The discriminator is pre-trained on a large-scale dataset of audio-image pairs, enabling it to assess synchronization discrepancies between lip motion and audio at the feature level. By incorporating the raw audio signal as supervision, the discriminator facilitates cross-modal fusion between visual and audio representations. During this stage, we optimize only the mouth prediction network and color parameters to ensure stable fine-tuning, i.e., the overall loss function is
\begin{equation}\label{equ}
\begin{aligned} 
        L = \mathcal{L}_{fu} + \lambda_3  \mathcal{L}_{lip}(x^{fu}, a),
\end{aligned}
\end{equation}
where $a$ is the input audio and $\lambda_3$ denotes the weight factor.

\section{Experiments}
\subsection{Experimental Setups}

\noindent \textbf{Datasets.} The experiments utilize high-definition speaking portrait videos sourced from publicly-released datasets that are well-established in prior works \cite{ye2023geneface,li2023efficient,tang2022real}. Each subject's dataset consists of several minutes of video of people speaking different languages, with a corresponding audio track, averaging approximately 6,500 frames per clip at 25 FPS.  To ensure a fair and reproducible comparison, we follow the preprocessing procedures of previous works: the raw videos are cropped to focus on the central portrait and then resized. The majority of the videos are set to a resolution of $512 \times 512$, except for the ``Obama'' video, which is resized to $450 \times 450$. All the videos are divided into training and test sets with a 10:1 ratio.\\

\begin{figure*}[t]
\centering
\includegraphics[width=\linewidth]{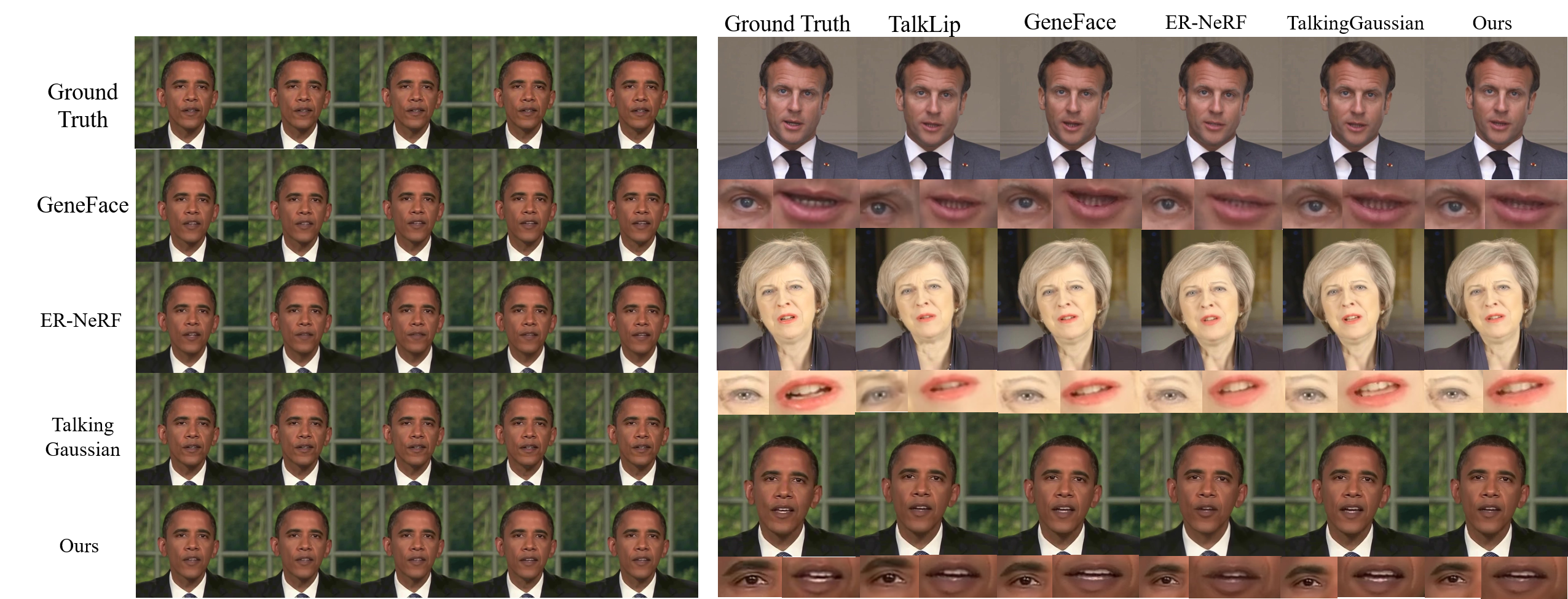}
\caption{\textbf{Qualitative comparison of talking head synthesis across different methods.} From top to bottom, rows show the ground truth and results produced by GeneFace, ER-NeRF, TalkingGaussian, and the proposed method. Close-up views highlight lip movements and fine facial details. The proposed method produces more accurate lip synchronization and more stable facial details, closely matching the ground truth. Zooming in is recommended for better visualization.}
\label{fig:visual_compare_with_other_methods}
\end{figure*}

\begin{table*}[t]
\caption{\textbf{The quantitative results of the talking head reconstruction.} The best performances are in boldface, and the underline represents the second-best performance.}
\centering
\renewcommand{\arraystretch}{1.4}
\resizebox{\linewidth}{!}{
    \setlength{\tabcolsep}{4mm}
    \centering
    \begin{tabular}{lcccccccc}
    \toprule
    Method & PSNR $\uparrow$ & LPIPS $\downarrow$ & FID $\downarrow$ & LMD $\downarrow$ & LSE-C $\uparrow$ & Time $\downarrow$ & FPS $\uparrow$ \\
    Ground Truth & N/A & 0 & 0 & 0 & 8.275 & - & - \\ \midrule
    TalkLip \cite{wang2023seeing} & 32.52 & 0.0782 & 18.500 & 5.861 & 5.947 & - & 3.41 \\
    DINet \cite{zhang2023dinet} & 31.65 & 0.0443 & 9.430 & 4.373 & \underline{6.565} & - & 23.74 \\
    \midrule
    AD-NeRF \cite{guo2021ad} & 26.73 & 0.1536 & 28.986 & 3.000 & 4.500 & 16.4h & 0.14 \\
    RAD-NeRF \cite{tang2022real} & 31.78 & 0.0778 & 8.657 & 2.912 & 5.522 & 5.2h & 53.87 \\ 
    GeneFace \cite{ye2023geneface}& 24.82 & 0.1178 & 21.708 & 4.286 & 5.195 & 12.3h & 7.79 \\
    ER-NeRF \cite{li2023efficient} & 32.52 & \underline{0.0334} & \underline{5.294} & \underline{2.814} & 5.775 & 3.1h & \underline{55.41} \\
    TalkingGaussian \cite{li2024talkinggaussian} & 32.40 & 0.0355 & 7.693 & 2.967 & 6.516 & \textbf{1h} & \textbf{90} \\
    PointTalk \cite{xie2025pointtalk} & \underline{32.77} & 0.0337 & 7.331 & 2.818 & \textbf{7.165} & \textbf{1h} & \textbf{90} \\
    \midrule
    FG-3DGS (Ours) & \textbf{33.06} & \textbf{0.0252} & \textbf{4.846} & \textbf{2.620} & 6.260 & \underline{2h} & \textbf{90} \\ \bottomrule 
    \end{tabular}
}
\label{tab:setting1}
\end{table*}

\noindent \textbf{Baseline Methods.} To validate the effectiveness of the proposed talking head model, the comparative methods included 2D-based, NeRF-based, and 3D Gaussian-based approaches. (i) 2D based methods include TalkLip~\cite{wang2023seeing} and DINet~\cite{zhang2023dinet}, (ii) NeRF-based methods include AD-NeRF~\cite{guo2021ad}, RAD-NeRF~\cite{tang2022real}, ER-NeRF~\cite{li2023efficient}, and GeneFace~\cite{ye2023geneface}, and (iii) 3D Gaussian-based methods TalkingGaussian~\cite{li2024talkinggaussian} and PointTalk~\cite{xie2025pointtalk}.

\noindent \textbf{Training Details.} For each subject, we train both the low-frequency and high-frequency regions for approximately 50,000 iterations. The training process is divided into two stages: 3,000 iterations for Static Gaussian Reconstruction and the remaining 47,000 for Frequency-Aware Modeling. Then, 10,000 more iterations are set for the joint Fusion \& Render stage. The hyperparameter settings for the first stage are inherited from 3DGS~\cite{kerbl3Dgaussians}. While the learning rate of the jointly emotion- and audio-driven network in the low-frequency region and the cross-modal attention block in the high-frequency region during the second and fusion stage is set to $1e-4$, while the learning rate of all other MLP networks is set to $1e-5$. All experiments are conducted on a single NVIDIA RTX 4090 GPU, and each person-specific model requires approximately two hours of training.

\subsection{Quantitative Evaluation}
For the quantitative evaluation, we conduct a comprehensive assessment of our proposed method across two distinct experimental settings: 1) the talking head reconstruction setting and 2) the cross-subject lip synchronization setting.
In the talking head reconstruction setting, we aim to evaluate the model's fidelity in reconstructing the appearance and motion of a specific subject driven by their original voice. In the cross-subject lip-synchronization setting, we focus on assessing the model's generalization performance when presented with out-of-distribution (OOD) audio inputs. For this purpose, we employ two standard audio clips, designated as Audio A and Audio B, sourced from public benchmarks established in prior work~\cite {suwajanakorn2017synthesizing}, following the settings of ER-NeRF~\cite{li2023efficient}. Notably, both audio clips originate from speakers other than the test target subjects, making generalization challenging. \\

\begin{figure}[t]
\begin{center}
\includegraphics[width=\linewidth]{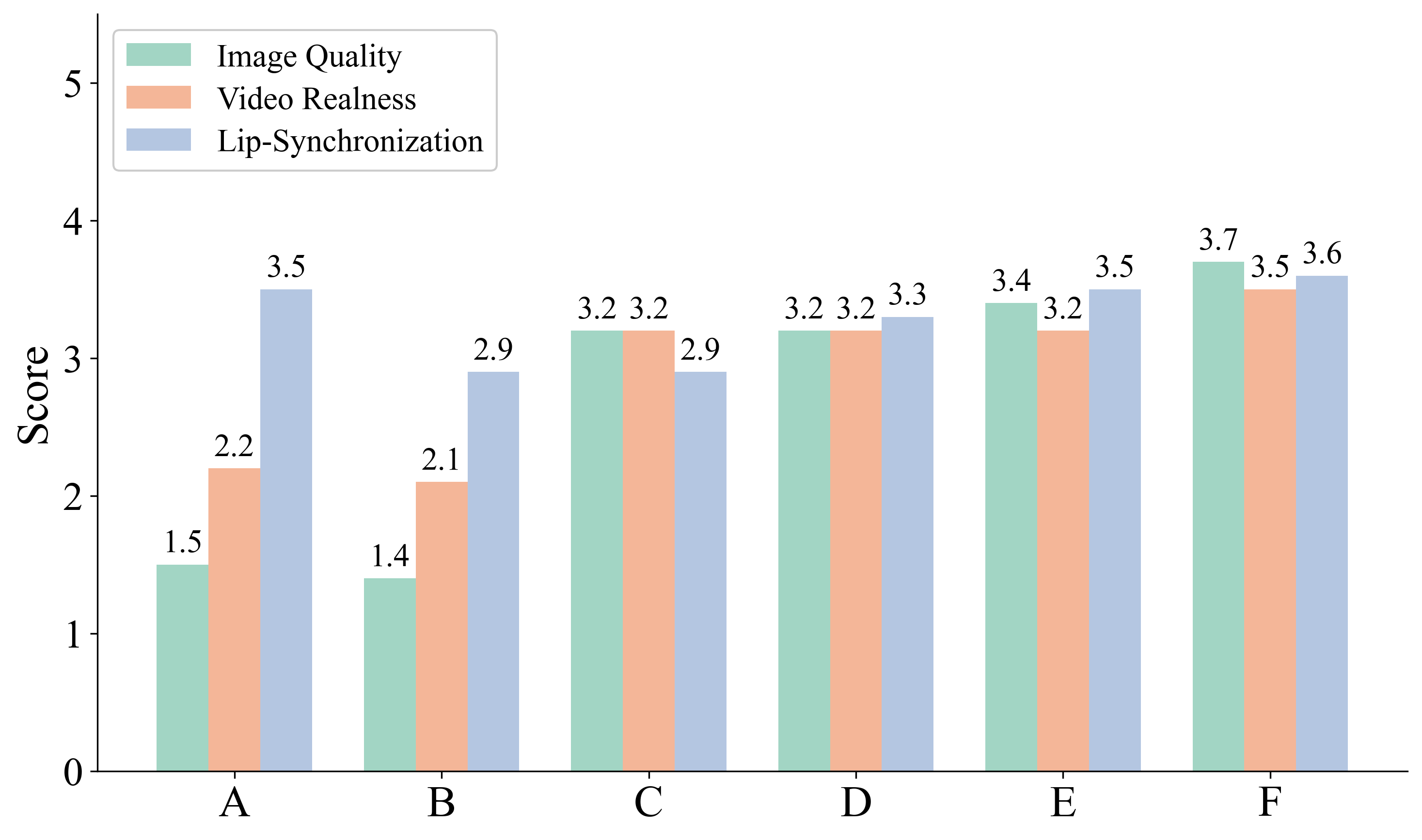}
\end{center}
\caption{\textbf{User study results.} Mean scores from 20 participants on a 5-point scale, where higher values indicate better performance. The evaluation covers image quality, video realism, and lip synchronization. Methods A--F correspond to TalkLip, DINet, ER-NeRF, GeneFace, TalkingGaussian, and the proposed method (FG-3DGS), respectively.}
\label{fig:the_user_study}
\end{figure}

\noindent \textbf{Metrics.} For the evaluation of the image quality, we utilize Peak Signal-to-Noise Ratio (PSNR) to measure overall reconstruction fidelity, the Learned Perceptual Image Patch Similarity (LPIPS)~\cite{zhang2018unreasonable} to evaluate the realism of high-frequency details, and the Frechet Inception Distance (FID)~\cite{heusel2017gans} to judge the image quality from the feature aspect. To evaluate lip synchronization and motion accuracy, we adopt the Landmark Distance (LMD)~\cite{chen2018lip}, which computes the direct Euclidean distance between generated and ground-truth lip landmarks. Furthermore, we leverage the pre-trained SyncNet~\cite{chung2016lip}~\cite{chung2016out_xxxx} to compute its lip synchronization estimation confidence score (LSE-C) and error distance (LSE-D), following the evaluation protocol of Wav2Lip~\cite{prajwal2020lip}. \\

\begin{table}[t]
\caption{\textbf{The quantitative results of the lip synchronization}. We utilize two different audio samples to drive the same subject. The boldface indicates the best performance, and the underline indicates the second-best.}
\centering
\renewcommand{\arraystretch}{1.4}
\resizebox{\linewidth}{!}{
\begin{tabular}{@{}lcccc@{}}
\toprule
 & \multicolumn{2}{c}{Audio A}       & \multicolumn{2}{c}{Audio B}       \\ \cmidrule(l){2-5} 
Method              & LSE-D $\downarrow$         & LSE-C $\uparrow$         & LSE-D $\downarrow$         & LSE-C $\uparrow$         \\ 
Ground Truth & 6.899           & 7.354         & 7.322             & 8.682         \\
\midrule

DINet \cite{zhang2023dinet}                                      & \textbf{8.503}          & 5.696          & \textbf{8.204}        & 5.113        \\
AD-NeRF \cite{guo2021ad}                                      & 14.432          & 1.274          & 13.896         & 1.877        \\
RAD-NeRF \cite{tang2022real}                                       & 11.639         & 1.941          & 11.082          & 3.135           \\
GeneFace \cite{ye2023geneface}                                    & 9.545         & 4.293          & 9.668          & 3.734          \\
ER-NeRF \cite{li2023efficient}                                       & 11.813           &  2.408        & 10.734        & 3.024        \\
TalkingGaussian \cite{li2024talkinggaussian}
&9.171            &  5.327        & 9.061        & 5.745        \\
\midrule
FG-3DGS (Ours)  & \underline{8.987}    & \textbf{6.197}    & \underline{9.039} & \textbf{6.317} \\ \bottomrule
\end{tabular}
}
\label{tab:setting2}
\end{table}

\noindent \textbf{Results.} Table \ref{tab:setting1} shows the quantitative results of the talking head reconstruction experiments. Among all the methods, our FG-3DGS ranks highest in most metrics. Due to the frequency-aware disentanglement and modeling strategy, our method achieves the highest PSNR and LPIPS, along with the lowest FID and LMD, demonstrating its ability to reconstruct image details while maintaining the highest render speed. Quantitative evaluation results demonstrate that the proposed FG-3DGS achieves the best overall performance on the talking head task studied in this paper.

\begin{figure}[t]
\begin{center}
\includegraphics[width=1\linewidth]{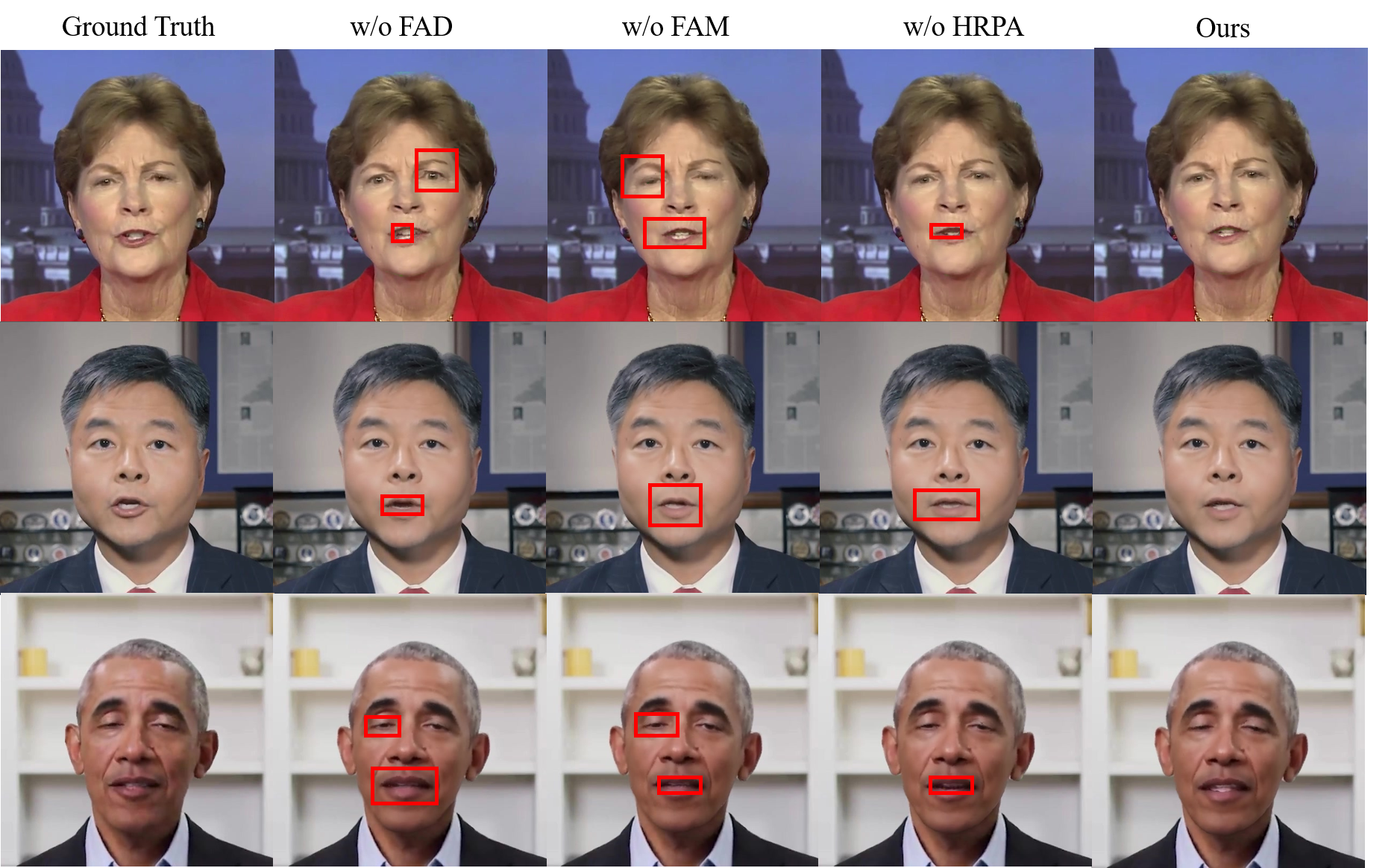}
\end{center}
\caption{\textbf{Qualitative results of the ablation study on the proposed components.} Each column shows the output obtained by removing one module (w/o FAD, w/o FAM, w/o HRPA) compared with the ground truth and the full model (Ours). Red bounding boxes highlight regions with visible artifacts and degradation, such as over-smoothing and inaccurate lip-synchronization.}
\label{fig:visual_results_of_ablation_study}
\end{figure}

Meanwhile, Table \ref{tab:setting2} reports the quantitative results of the cross-subject lip synchronization by different driven audios. Our proposed method achieves the best LSE-C performance among all competitors, indicating superior lip-synchronization confidence and semantic alignment. Notably, while DINet (a 2D-based method) yields a slightly lower LSE-D, our approach significantly outperforms all 3D-based and NeRF-based frameworks (e.g., GeneFace, TalkingGaussian) by a large margin in both metrics. The competitive LSE-D, coupled with the state-of-the-art LSE-C, demonstrates that our method effectively bridges the gap between 3D controllability and high-fidelity lip synchronization, maintaining high-quality lip alignment while overcoming the typical limitations of 3D facial animation in capturing precise lip dynamics.

\subsection{Qualitative Evaluation}
To more intuitively evaluate the performance of the generated talking heads, we present keyframes from our talking head reconstruction experiments. We mainly focus on the portrait's reconstruction details and lip-synchronization effect. As shown in Fig.~\ref{fig:visual_compare_with_other_methods}, when other methods fail to capture the detailed appearance of the mouth and eye region, FG-3DGS generates a high-quality talking face. Moreover, our method maintains high fidelity and lip-synchronization accuracy in sequential frames, as shown in the left part of Fig.~\ref{fig:visual_compare_with_other_methods}.

To complement our quantitative metrics, we conduct a user study. We recruited 20 participants and presented them with 36 talking-head videos generated by our method and five others. Participants were asked to rate each video on a 5-point scale according to three distinct criteria: (1) Image Quality (clarity, texture, and fidelity), (2) Video Realness (naturalness of motion and expressions), and (3) Lip-Synchronization (the consistency between lip and the audio). The scores are summarized in Fig.~\ref{fig:the_user_study}, indicating that FG-3DGS achieved the highest ratings across all three categories, outperforming all baseline methods.

\subsection{Ablation Study and Discussion}

This section primarily analyzes the components and configurations of the experimental design, including the proposed frequency decomposition strategy, the audio extractor, the selection of motion offsets, and a discussion of hyperparameter sensitivity. The details are as follows.\\

\noindent \textbf{Proposed Components.}
We conduct an ablation study across different settings to demonstrate that our fine-grained strategy makes a significant contribution to high-quality talking head generation. The results in Table \ref{tab:ablation_study}.
show that without Frequency-Aware Disentanglement (FAD), the total reconstruction metrics decline slightly, indicating that the fine-grained strategy helps control facial details. The degradation in LPIPS and PSNR metrics is attributed to the absence of Frequency-Aware Modeling (FAM), which reduces the degree of fusion between audio and spatial features, thereby lowering the overall quality of the generated image. The lack of the High-frequency Refined Post-rendering Alignment (HRPA) results in reduced synchronization between the lips and the audio. These results demonstrate the efficacy of our proposed components. Meanwhile, the visual results in Fig.~\ref{fig:visual_results_of_ablation_study} further validate the effectiveness of the proposed components.\\

\begin{table}[tb]
\caption{\textbf{Ablation study of proposed components.} Best performance is shown in bold.}
\centering
\small
\renewcommand{\arraystretch}{1.3}

    \setlength{\tabcolsep}{2mm}
    \begin{tabular}{lcccccc}
    \toprule
    Setting  & PSNR$\uparrow$ & LPIPS$\downarrow$  & FID$\downarrow$ & LMD$\downarrow$  & LSE-C$\uparrow$\\ \midrule
    w/o FAD & 32.93  & 0.0275 & 5.396    & 2.719  & 6.111    \\
    w/o FAM & 32.86  & 0.0289  &7.150   & 2.854  & 5.965   \\
    w/o HRPA & 32.89 & 0.0261   &6.715   & 2.773 & 5.950    \\
    FG-3DGS & \textbf{33.06} & \textbf{0.0252} & \textbf{4.846} & \textbf{2.620} & \textbf{6.260}   \\
    \bottomrule
    \end{tabular}
\setlength{\abovecaptionskip}{0cm}
\label{tab:ablation_study}
\end{table}

\noindent \textbf{Audio Extractor.} To ensure a fair comparison with existing state-of-the-art talking head methods, we primarily use Deepspeech~\cite{hannun2014deep} as the audio feature extractor. To further demonstrate the generalizability and performance upper bound of our proposed model, we also evaluate its performance when integrated with more sophisticated extractors. As quantitatively shown in Table~\ref{tab:audio_extractor}, substituting the baseline with HuBERT~\cite{hsu2021hubert} or Wav2Vec 2.0~\cite{baevski2020wav2vec} leads to consistent performance improvements. This outcome indicates that our model can effectively leverage high-quality phonetic representations to facilitate finer lip-audio synchronization and more realistic speech-driven dynamics.\\

\begin{table}[tb]
\caption{\textbf{Quantitative metrics for adopting different audio extractors in the talking head reconstruction setting.}}
\centering
\renewcommand{\arraystretch}{1.5}
\resizebox{1\linewidth}{!}{
    \setlength{\tabcolsep}{2mm}
    \begin{tabular}{lcccccc}
    \toprule
    Extractor  & PSNR$\uparrow$ & LPIPS$\downarrow$  & FID$\downarrow$ & LMD$\downarrow$  & LSE-C$\uparrow$\\ \midrule
    Deepspeech & \textbf{33.06} & 0.0252 & \textbf{4.846} & 2.645 & 6.260 &\\
    HuBERT & 33.05 & \textbf{0.0251} & 4.903 & 2.643 & \textbf{6.401} &\\
    Wav2Vec 2.0 & 33.03 & 0.0254 &  4.961& \textbf{2.642} & 6.269 & \\
    \bottomrule
    \end{tabular}
}

\setlength{\abovecaptionskip}{0cm}
\label{tab:audio_extractor}
\end{table}

\begin{figure}[tb]
\begin{center}
\includegraphics[width=1\linewidth]{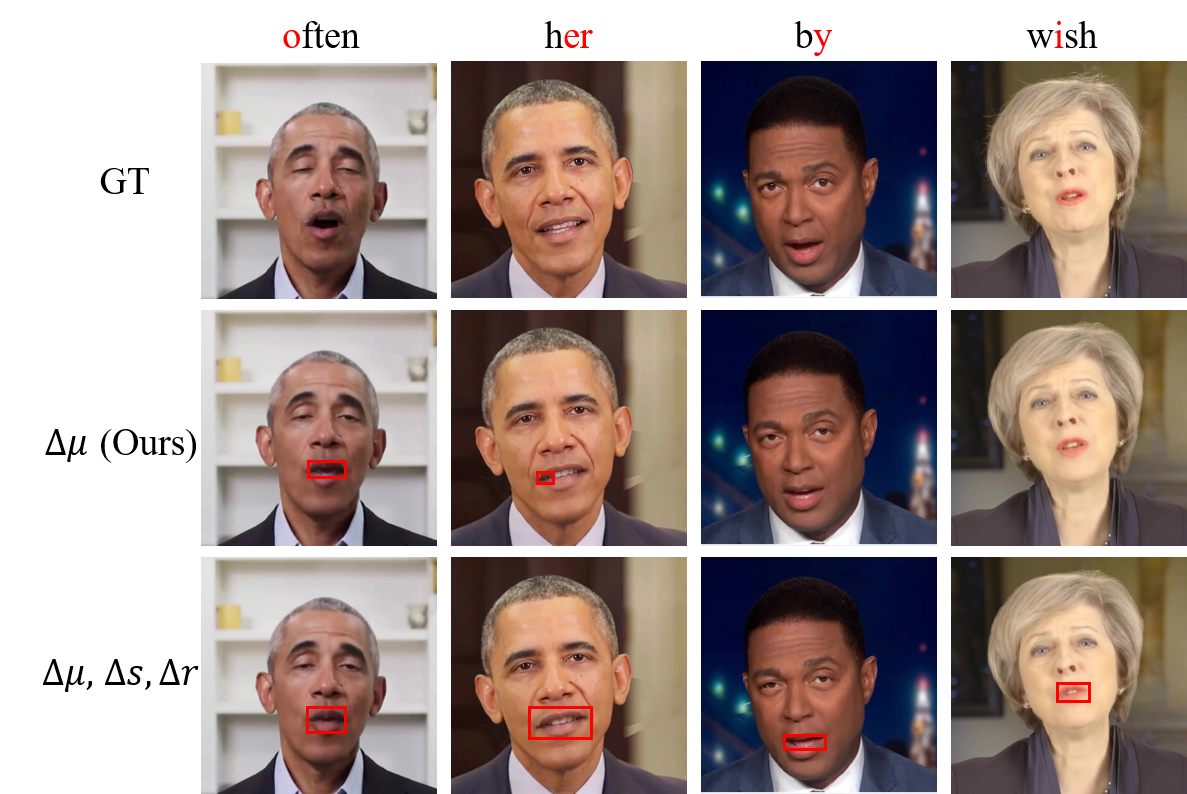}
\end{center}
\caption{\textbf{Qualitative comparison of visual results under different motion-offset selection strategies.} Red bounding boxes highlight regions exhibiting over-smoothing artifacts and inaccurate lip-synchronization. Results are shown for the ground truth (GT), the proposed mean offset only ($\Delta \mu$), and the full configuration incorporating mean, scale, and residual offsets ($\Delta \mu$, $\Delta s$, $\Delta r$).}
\label{fig:visual of offset}
\end{figure}

\noindent \textbf{Motion Offset.} We extensively conduct experiments to explore how various selections of motion parameter subsets influence the final synthesis quality. Fig.~\ref{fig:visual of offset} presents a visualization of the comparative results. Our empirical findings suggest that when simultaneously predicting the set of Gaussian attributes $\{\Delta \mu, \Delta r, \Delta s\}$ in the high-frequency zones, the model tends to suffer from optimization instability. This over-parameterization often leads to perceptible over-smoothing of fine details and significantly reduces lip-synchronization accuracy. Consequently, we select only the position offset $\Delta \mu$ as the target for our motion prediction network in high-frequency regions to ensure robust motion synthesis.\\

\begin{table}[tb]
\caption{\textbf{Quantitative metrics under varying hyperparameters.} The default settings are $\lambda_1=0.20$, $\lambda_2=0.50$, and $\lambda_3=0.03$.}
\centering
\small 

\setlength{\tabcolsep}{4.5pt} 
\renewcommand{\arraystretch}{1.0}

\begin{tabular}{lccccc}
\toprule
Setting & PSNR $\uparrow$ & LPIPS $\downarrow$ & FID $\downarrow$ & LMD $\downarrow$ & LSE-C $\uparrow$ \\ 
\midrule
$\lambda_1=0.10$ & 33.03 & 0.0263 & 5.052 & 2.685 & 6.099 \\
$\lambda_1=0.15$ & \textbf{33.08} & 0.0256 & 4.946 & 2.623 & 6.233 \\
$\lambda_1=0.20$  & 33.06 & 0.0252 & \textbf{4.846} & \textbf{2.620} & \textbf{6.260} \\
$\lambda_1=0.30$ & 32.99 & \textbf{0.0251} & 5.101 & 2.641 & 5.987 \\
\midrule
$\lambda_2=0.40$ & 32.94 & 0.0266 & 5.213 & 2.653 & 6.189 \\
$\lambda_2=0.45$ & 32.96 & 0.0261 & 5.058 & 2.644 & 6.210 \\
$\lambda_2=0.50$  & 33.06 & 0.0252 & \textbf{4.846} & \textbf{2.620} & \textbf{6.260} \\
$\lambda_2=0.60$ & \textbf{33.10} & \textbf{0.0251} & 5.106 & 2.671 & 6.015 \\
\midrule
$\lambda_3=0.01$ & 33.01 & 0.0255 & 4.913 & 2.645 & 6.211 \\
$\lambda_3=0.02$ & 32.99 & \textbf{0.0249} & 5.718 & 2.680 & 6.125 \\
$\lambda_3=0.03$  & \textbf{33.06} & 0.0252 & \textbf{4.846} & \textbf{2.620} & \textbf{6.260} \\
$\lambda_3=0.04$ & 33.03 & 0.0254 & 4.905 & 2.633 & 6.116 \\
\bottomrule
\end{tabular}
\label{tab:ablation_hyperparameters}
\end{table}

\noindent \textbf{Hyperparameter Sensitivity.} During the fusion stage, we use the $L_1$ loss, D-SSIM loss $\mathcal{L}_{DS}$, LPIPS loss $\mathcal{L}_{LP}$, and a task-specific lip-synchronization loss $\mathcal{L}_{lip}$ derived from a pretrained lip-discriminator.  The relative contributions within the total loss function $\mathcal{L}_{fu}$ are empirically set as follows: $\lambda_1=0.2$ for structural constraints, $\lambda_2=0.5$ for perceptual alignment, and $\lambda_3=0.03$ for synchronization. We conduct a series of comprehensive quantitative experiments to identify the most effective hyperparameter configurations across different training phases. As detailed in Table~\ref{tab:ablation_hyperparameters}.  (1) The sensitivity analysis for $\lambda_{3}$ in the lip-synchronization loss demonstrates that $\lambda_{3}=0.03$ achieves the best results of PSNR, LPIPS, FID, and LMD. Deviating from this value often degrades either visual quality or speech alignment accuracy, making it the final parameter. (2) The coefficients $\lambda_{1}$ and $\lambda_{2}$ serve as regulators for reconstruction fidelity. They measure the magnitude of structural and textural error between the generated talking head image and the ground-truth frame, ensuring the synthesized results remain anchored to the target identity's appearance.

\section{Conclusion}
This paper presents FG-3DGS, a novel framework for fine-grained audio-driven 3D talking head generation. FG-3DGS addresses the challenge of precise facial control—critical for avoiding the uncanny valley effect—by identifying that prior methods often overlook the distinct motion dynamics of different facial regions. To overcome this, the framework disentangles low- and high-frequency facial motions, enhancing high-frequency regions via a high-frequency-refined post-rendering alignment. With these specialized components, FG-3DGS consistently outperforms recent state-of-the-art methods across various settings, demonstrating the effectiveness of frequency-aware disentanglement for fine-grained, realistic 3D talking head generation.

\section{Supplementary Video}
We provide a set of video results to demonstrate our method's performance across both talking-head reconstruction and cross-subject lip-synchronization settings. The corresponding files are organized into the ``reconstruction'' and ``cross\_subject'' folders, respectively.

\bibliography{ref}
\bibliographystyle{IEEEtran}

\end{document}